\documentclass[10pt,twocolumn,letterpaper]{article}

\usepackage[pagenumbers]{cvpr} 

\usepackage{url}            
\usepackage{booktabs}       
\usepackage{amsfonts}       
\usepackage{nicefrac}       
\usepackage{microtype}      
\usepackage{xcolor}         
\usepackage{graphicx}
\usepackage{wasysym}
\usepackage{cases}
\usepackage{multirow}
\usepackage{color}

\usepackage{amsmath}

\definecolor{cvprblue}{rgb}{0.21,0.49,0.74}
\usepackage[pagebackref,breaklinks,colorlinks,allcolors=cvprblue]{hyperref}

\def\paperID{6536} 
\def\confName{CVPR}
\def\confYear{2026}

\title{Unbiased Dynamic Multimodal Fusion}

\author{
Shicai Wei$^{}$, Kaijie Zhang$^{}$, Luyi Chen$^{}$, Tao He$^{}$, Guiduo Duan$^{*}$\\
$^{}$University of Electronic Science and Technology of China\\
{\tt\small shicaiwei@uestc.edu.cn,duanguiduo@163.com}
}

\begin{document}
\maketitle
\begin{abstract}
Traditional multimodal methods often assume static modality quality, which limits their adaptability in dynamic real-world scenarios. Thus, dynamical multimodal methods are proposed to assess modality quality and adjust their contribution accordingly. However, they typically rely on empirical metrics, failing to measure the modality quality when noise levels are extremely low or high. Moreover, existing methods usually assume that the initial contribution of each modality is the same, neglecting the intrinsic modality dependency bias. As a result, the modality hard to learn would be doubly penalized, and the performance of dynamical fusion could be inferior to that of static fusion. To address these challenges, we propose the Unbiased Dynamic Multimodal Learning (UDML) framework. Specifically, we introduce a noise-aware uncertainty estimator that adds controlled noise to the modality data and predicts its intensity from the modality feature. This forces the model to learn a clear correspondence between feature corruption and noise level, allowing accurate uncertainty measure across both low- and high-noise conditions. Furthermore, we quantify the inherent modality reliance bias within multimodal networks via modality dropout and incorporate it into the weighting mechanism. This eliminates the dual suppression effect on the hard-to-learn modality. Extensive experiments across diverse multimodal benchmark tasks validate the effectiveness, versatility, and generalizability of the proposed UDML.  The code is available at \url{https://github.com/shicaiwei123/UDML}.



\end{abstract}




\section{Introduction}

Multimodal learning has achieved significant progress across a wide range of vision tasks, including classification~\cite{mm_cf1,mm_cf2,wei2025boosting}, object detection~\cite{mm_detection1,mm_detection2,mm_detection3}, and segmentation~\cite{rgbd_seg1,rgbd_seg2,rgbd_seg3}. Most existing multimodal learning methods assume that the quality of each modality is static, with one modality consistently stronger than the others. However, this assumption often fails in real-world scenarios. For example, the RGB modality typically provides richer information than the IR modality during daytime but suffers substantial reliability degradation at night, whereas the IR modality often outperforms RGB under low-light conditions~\cite{li2018densefuse}. Therefore, it is essential to develop multimodal learning methods capable of handling dynamic data quality.

\begin{figure}[t]
\centering
\includegraphics[width=0.51\textwidth]{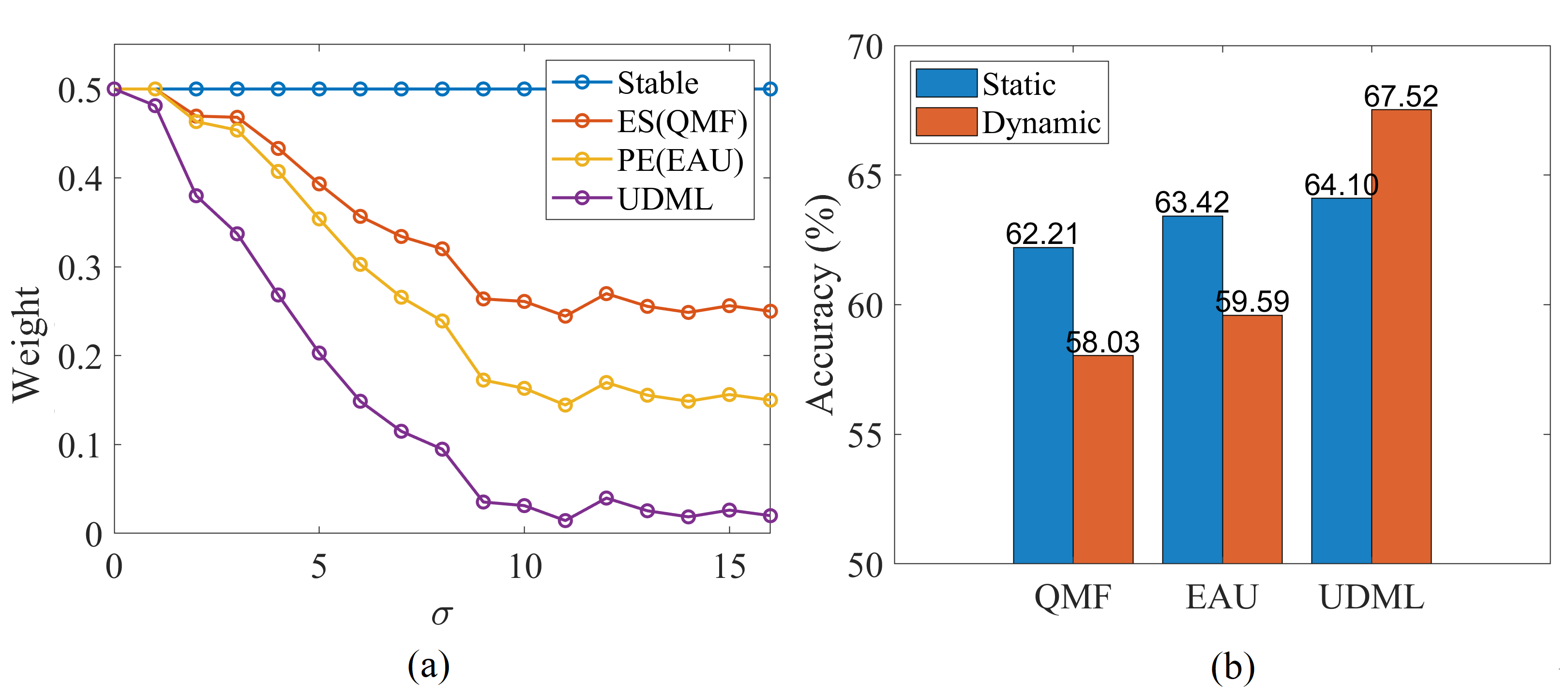} 
\caption{ Visualization of dynamic multimodal methods for audio-visual classification on the CREMA-D dataset. (a) Visual weighting coefficients obtained using different uncertainty estimation methods, such as energy score (ES)~\cite{qmf} and probabilistic embedding (PE)~\cite{eau}, and the proposed UDML, as varying levels of noise ($\sigma$) are injected into the visual modality.  (b) Performance Comparison of different methods under static and dynamic weighting when noise ($\sigma=5$) is injected into the visual modality.}
\label{fig:dynamic_fusion}
\end{figure}

To address the dynamic quality of multimodal data, numerous methods have been proposed to assess modality reliability and adjust their contributions accordingly. These methods can be broadly categorized into two paradigms: prior-based methods~\cite{late2,li2018densefuse} and uncertainty-based methods~\cite{tmc,eau,ldd,qmf}. Prior-based methods estimate modality quality using human knowledge or experience. For instance, LiDAR sensors may be prioritized at night while cameras are favored in daylight in autonomous driving. While intuitive, such approaches often lack generalizability, particularly in complex or unforeseen environments. Thus, uncertainty-based methods have recently gained attention as a more principled and generalizable solution. By leveraging probabilistic modeling or information-theoretic measures, these methods dynamically adjust modality contributions, offering a more robust foundation for reliable multimodal integration.

Despite advances in dynamic multimodal learning, existing approaches still suffer from two major limitations. First, most existing uncertainty estimation methods rely on empirical metrics, such as energy score~\cite{qmf} and probabilistic embedding~\cite{eau,ldd}, failing to assess the modality quality when noise levels are either too low or too high. Taking the widely used probabilistic embedding (PE) as an example (Fig.~\ref{fig:dynamic_fusion}(a)), we observe that PE fails to detect low-intensity noise($\sigma<4$), leading to rigid modality weighting. Besides, even under extreme corruption ($\sigma>10$), PE still assigns considerable weight to the corrupted modality rather than disregarding it. Second, existing methods generally assume that the initial contributions of all modalities are the same, overlooking modality dependence biases. In practice, multimodal models tend to rely more on easy-to-learn modalities to make decisions~\cite{ogm,balance3}. As a result, the modality hard to learn would be doubly penalized due to optimization bias and high uncertainty. Consequently, as shown in Fig~\ref {fig:dynamic_fusion}(b), the performance of dynamic fusion could be inferior to that of the stable fusion.


To this end, we propose an Unbiased Dynamic Multimodal Learning (UDML), a general framework to assist multimodal learning with dynamical data quality. UDML consists of two key components: a noise-aware uncertainty estimator and a modality-dependency calculator. The noise-aware uncertainty estimator adds controlled noise to the modality data and predicts its intensity from the modality feature, achieving accurate estimation across both low- and high-noise regimes. To further ensure robustness to unseen noise types, we introduce the probabilistic representation technique that maps each modality into a distribution to decouple noise from semantic content. Specifically, the mean encodes semantic information, and the variance reflects noise characteristics. The estimator then derives noise intensity directly from the variance.  The modality-dependency calculator employs modality dropout to quantify the output’s inherent reliance on each modality. This dependency measure is then used to recalibrate modality weights, balancing their contributions and mitigating dual-suppression effects. Notably, UDML is architecture-agnostic, as all components operate solely on modality representations, ensuring generalizability across diverse models and fusion methods. Finally, we introduce a progressive optimization strategy that enables the simultaneous learning of multimodal representations, noise estimation, and the primary task within a standard training schedule. Extensive experiments on multiple tasks and datasets verify that UDML consistently enhances multimodal performance, demonstrating its robustness and versatility.


\begin{itemize}
    \item We reveal the bias of existing uncertainty-based estimators, which are insensitive to slight degradation and assign non-negligible weights to heavily corrupted modalities. This limits the robustness of dynamic fusion.
    \item We reveal the dual suppression effect in existing dynamic multimodal learning, in which the modality hard to learn is doubly penalized due to optimization bias and high uncertainty. This could lead to the dynamic fusion underperforming the static fusion.
    \item We propose Unbiased Dynamic Multimodal Learning (UDML), an architecture-agnostic framework explicitly addressing both quality estimation bias and dual-suppression bias in dynamic fusion, collectively ensuring robust dynamic multimodal learning.
    \item Extensive experiments on diverse multimodal benchmarks, demonstrating that UDML consistently improves performance across various tasks and settings.
\end{itemize}

\section{Related Work}

\subsection{Dynamic Multimodal Learning}
Researchers have proposed a series of dynamic multimodal fusion algorithms, which can be categorized into two classes: prior-based methods~\cite{late2,li2018densefuse} and uncertainty-based methods~\cite{eau,tmc,qmf,ldd}.


Prior-based fusion methods allocate modality weights based on human knowledge. For example, the RGB modality typically contains more information than the infrared (IR) modality under normal lighting conditions. However, this relationship can reverse in low-light scenarios, where infrared imaging becomes more reliable. To address this, Guan et al. ~\cite{late2} introduce an illumination-aware fusion module that dynamically adjusts modality contributions based on the scene's lighting intensity. In addition to environmental factors, intrinsic properties of network features can also inform fusion decisions. For instance, Li et al. ~\cite{li2018densefuse} leverage the scaling factors in batch normalization as a feature selection metric to adjust the contribution of different modalities.



Uncertainty-based fusion methods adjust modality contributions based on prediction uncertainty, offering a more theoretically grounded approach compared to prior-based methods. A commonly used uncertainty metric is predictive entropy computed from the classifier logits~\cite{tmc,arl,mmanet,wei2024scaled}. However, since incorrect predictions may still yield high-confidence scores, relying solely on classifier outputs may lead to overconfidence. Thus, feature-space modeling with multivariate Gaussian distributions has been proposed to more accurately capture modality uncertainty by analyzing feature variance. This approach has seen extensive application in emotion recognition ~\cite{wei2024robust,tellamekala2023coldfusion,ldd} and image-text classification ~\cite{eau,qmf}



Despite their success, most existing uncertainty estimation methods rely on empirical heuristics, failing to assess the modality quality when noise levels are either too low or too high. More importantly, current studies assume that the initial contribution of each modality is equal. This overlooks the modality dependency imbalances induced by optimization bias. Consequently, the performance of dynamic fusion could be inferior to that of stable fusion.

\subsection{Imbalanced Multimodal Learning}

Recent studies pointed out that most multimodal learning methods fail to enhance performance significantly, even with more information~\cite{ogm,umt,wh,pmr,mmcosine,MLA}. Wang~\etal~\cite{wh} observed that different modalities exhibit varying convergence rates, leading to multimodal models that fail to surpass their unimodal counterparts. Peng~\etal~\cite{ogm} further showed that the modality with superior performance tends to dominate the optimization process, leading to inadequate feature learning in weaker modalities. To this end, various methods have been developed to enhance the conventional multimodal learning framework and can be roughly categorized into two types: gradient modulation and alternating optimization. Gradient modulation~\cite{ogm,pmr,mmcosine} aims to enlarge the gradient of weaker modality in multimodal learning, balancing the optimization of different modality encoders. Alternating optimization methods~\cite{mmpareto,diagnosing,MLA,reconboost} transform the conventional joint multimodal learning process into an alternating unimodal learning process to minimize inter-modality interference directly.

However, these methods address only the underoptimization problem of modality encoders during the training stage and do not correct the model's dependence bias on each modality during inference.

\begin{figure*}[t]
	\centering
	\includegraphics[width=1.0\textwidth]{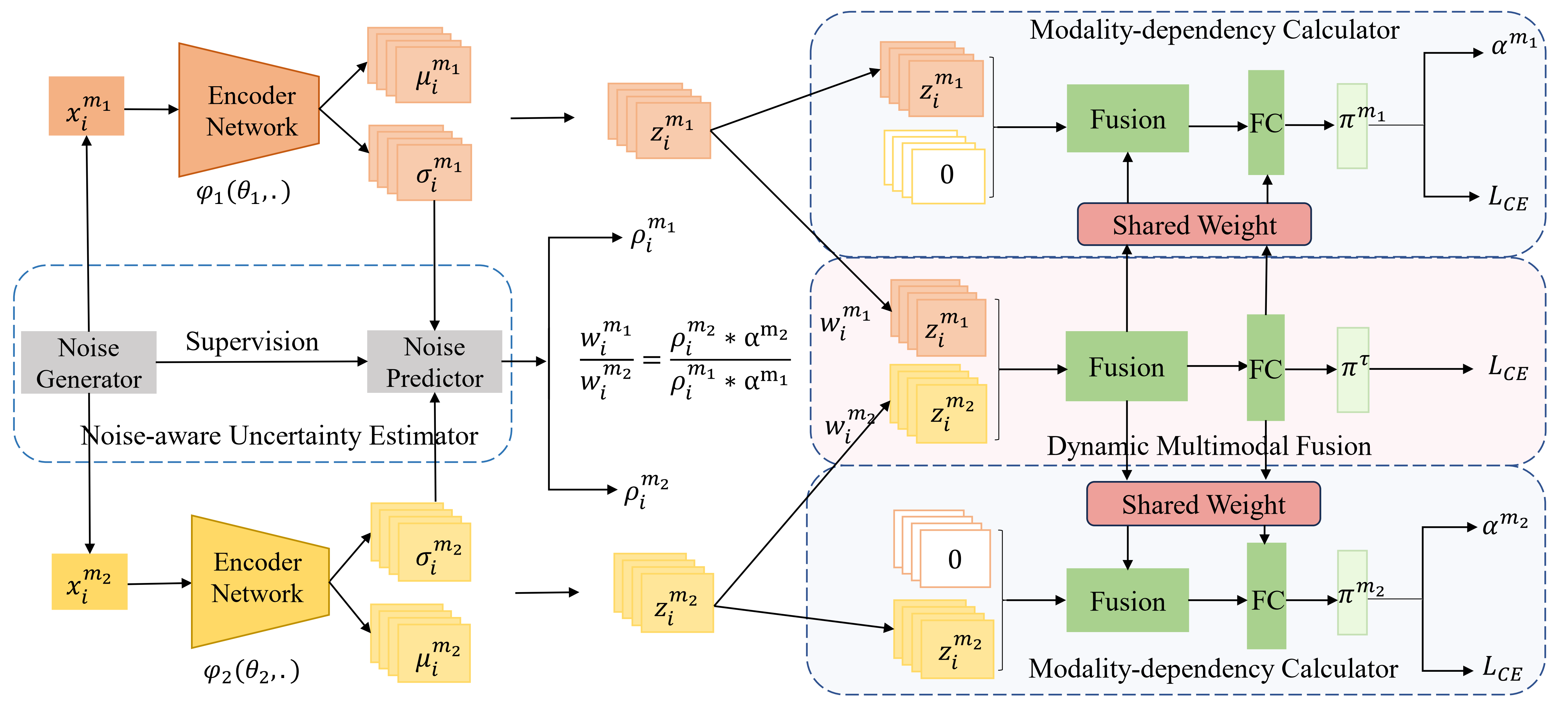}
    
	\caption{The framework of the unbiased dynamic multimodal fusion. It consists of two parts: 1)  noise-aware uncertainty estimator,  which measures the modality quality; 2) modality-dependency calculator, which quantifies the model's dependency on each modality.}
	\label{ogt}
\end{figure*}

\section{Methods}

\subsection{Re-analyze the Dynamic  Multimodal Learning}

\textbf{Notation}. In a general setting, we consider two input modalities denoted as \( m_1 \) and \( m_2 \). The dataset is represented as \( \mathcal{D} = \{x_i^{m_1}, x_i^{m_2}, y_i\}_{i=1,2,\dots,N} \), where \( y \in \{1,2,\dots,K\} \) denotes the class labels and \( K \) is the total number of classes. We then employ two encoders \( \varphi_{1}(\boldsymbol{\theta}_{1}, \cdot) \) and \( \varphi_{2}(\boldsymbol{\theta}_{2}, \cdot) \) to extract features, where \( \boldsymbol{\theta}_1 \) and \( \boldsymbol{\theta}_2 \) represent the parameters of each respective encoder. The feature representations are given by \( \boldsymbol{z}_i^{m_1} = \varphi_1(\boldsymbol{\theta}_1, x_i^{m_1}) \) and \( \boldsymbol{z}_i^{m_2} = \varphi_2(\boldsymbol{\theta}_2, x_i^{m_2}) \). The representations from the two encoders are typically fused via a specific fusion method, which is a common practice in multimodal learning \cite{MARS, tensor-1}. We denote the fusion module as \( \phi_{\tau}(\boldsymbol{\theta}_{\tau}, \cdot) \), where \( \boldsymbol{\theta}_{\tau} \) represents the parameters of this module. The final classification is performed by a linear classifier parameterized by \( \boldsymbol{W} \in \mathbb{R}^{M \times (d_1 + d_2)} \) and \( \boldsymbol{b} \in \mathbb{R}^M \), and the model output for an input \( x_i \) can be expressed as follows:
\begin{numcases}{}
	f({x}_{i})=\boldsymbol{W} \boldsymbol{z_i}^{\tau}+\boldsymbol{b} \\
	\label{zf}
	\boldsymbol{z_i}^{\tau}= \phi_{{\tau}}(\boldsymbol{\theta_{{\tau}}},\boldsymbol{z}_i^{m_1};\boldsymbol{z}_i^{m_2}).
\end{numcases}


\textbf{Dynamic Multimodal Learning (DML)}. DML considers the problem that the capacity of $z_{i}^{m_1}$ and $z_{i}^{m_2}$ varies with the input samples. Consequently, the multimodal model should adjust the decision dependency on each modality dynamically according to their representation capacity. Thus, the fusion representation of DML can be written as follows,

\begin{equation}
    {z_i}^{\tau-dml}=\varphi_{\tau}(\theta_{\tau},M*{w_i}^{m_1}*{z_i}^{m_1},M*{w_i}^{m_2}*{z_i}^{m_2}),  
\end{equation} where ${w_i}^{m_1}+{w_i}^{m_2}=1$; $M$ is the number of modalities, which is used to compensate for the amplitude suppression from the weight. Here, $\frac{{w_i}^{m_1}}{{w_i}^{m_2}}$ is  proportional to the representation capacity of modality $m_1$ and $m_2$, assigning the low-capacity modality a lower attention.

Therefore, the accurate measure of modality capacity is the key of DML. Most existing methods quantify the modality capacity via statistical uncertainty measures $s(.)$,such as energy fraction~\cite{qmf} and probabilistic embedding~\cite{pe-face-1},
\begin{equation}
{w_i}^{m_1} = g(\frac{1}{s({z_i}^{m_1})}),
\end{equation}
where $g(.)$ is the normalization operator. Lower uncertainty means higher capacity~\cite{eau,qmf}.


\textbf{Limitation Analysis}.  Existing methods inherently assume a stable and monotonic relationship between $s(z)$ and modality uncertainty. In practice, this assumption does not hold. Under low noise, $s(z)$ is insensitive to subtle quality differences, leading to nearly fixed weights. Under high noise, representation collapse distorts $s(z)$, producing inflated or misleading uncertainty estimates. Taking the widely used probabilistic embedding (PE) as an example (Fig.~\ref{fig:dynamic_fusion}(a)), PE fails to detect low-intensity noise ($\sigma<4$), resulting in rigid modality weighting. More critically, even under extreme corruption ($\sigma>10$), PE still assigns considerable weight to the corrupted modality rather than disregarding it.


Besides, existing methods generally assume that the initial contributions of all modalities are identical, overlooking the intrinsic dependency bias. Specifically, we denote the intrinsic dependency of modality $m$ as $\alpha^{m}$, satisfying $\sum_{m} \alpha^{m}=M$. We can rewrite ${z_i}^{\tau-dml}$  as follows:

\begin{equation}
    {z_i}^{\tau-dml}=\varphi_{\tau}(\theta_{\tau},M*{w_i}^{m_1}*\alpha^{m_1}*{z_i}^{m_1},M*{w_i}^{m_2}*\alpha^{m_2}{z_i}^{m_2}).
\end{equation}

Ideally, $\alpha^{m_1}=\alpha^{m_2}=1$ in a balanced multimodal system. However, multimodal models tend to rely more on easy-to-learn modalities to make decisions~\cite{ogm,balance3}, leading to an unbalanced dependency. Taking the audio-visual task as an example, assuming that audio modality is $m_1$  and visual modality is $m_2$, we can obtain the following inequality:
\begin{equation}
{\alpha}^{m_1}>1>{\alpha}^{m_2}.
\end{equation}


This is because audio modality is easier to learn than the visual one~\cite{ogm,balance3}. This will cause the dual suppression problem when the visual modality with high uncertainty. The hard-to-learn visual modality is first suppressed by optimization bias (low $\alpha^{m_2}$), and then further down-weighted by uncertainty-based reweighting (low ${w_i}^{m_2}$). Consequently, as illustrated in Fig.~\ref{fig:dynamic_fusion}(b), the performance of dynamic fusion could be inferior to that of static fusion.

\subsection{Unbiased Dynamic  Multimodal Learning}
Accordingly, conventional dynamic multimodal learning suffers from the biases of uncertainty estimation and dual suppression. To overcome these challenges, we propose an unbiased dynamic multimodal learning framework. As shown in Fig~\ref{ogt}, it consists of two key components: a noise-aware uncertainty estimator to quantify modality uncertainty (i.e., $ {{\rho}_i}^{m_1}$ and ${\rho_i}^{m_2}$) as well as a modality-dependency calculator to quantify modality effect on outputs(i.e., $ {\alpha}^{m_1}$ and ${\alpha}^{m_2}$). Then we can get the unbiased weight ${w_i}^{m_1}$,${w_i}^{m_2}$ as follows:

\begin{equation}
    \frac{{w_i}^{m_1}}{{w_i}^{m_2}}=\frac{\frac{1}{{\rho_i}^{m_1}*{\alpha}^{m_1}}}{\frac{1}{{\rho_i}^{m_2}*{\alpha}^{m_2}}}.
\end{equation}
Beyond conventional dynamic multimodal learning, the unbiased weight considers the modality-dependency bias of each modality, avoiding the problem of dual suppression. Finally, since the joint training of multimodal representations, uncertainty estimation, and the dependency calculator may be unstable, we design a progressive optimization strategy to ensure convergence.


\subsubsection{ Noise-aware Uncertainty Estimator}

Existing uncertainty-based methods often fail to assess the modality quality when noise levels are either too low or too high. To this end, we introduce a noise-aware uncertainty estimator that learns to predict the intrinsic noise intensity of each modality through controlled perturbations.

Take the modality $m_1$ as an example, let $p(\sigma)$ denote a discrete set of noise levels used for perturbation. For each sample ${x_i}^{m_1}$ and noise level $\sigma\sim p(\sigma)$ we draw perturbations $\epsilon\sim\mathcal{N}(0,\sigma^2 I)$ and supervise the estimator $E(\cdot)$ to predict the scalar intensity $\sigma$. The training objective is

\begin{equation}
\mathcal{L}_{\mathrm{est}} 
=\big\| E({x_i}^{m_1}+\epsilon_{i,k})-\sigma_k\big\|_2^2,
\end{equation}
where $\sigma_k\sim p(\sigma)$, and $\epsilon_{i,k}\sim\mathcal{N}(0,\sigma_k^2 I)$.

This formulation forces the model to learn a clear correspondence between feature corruption and noise level, allowing accurate uncertainty measure across both low- and high-noise conditions.

To avoid overfitting to the injected perturbation and improve robustness to unseen noise types, we do not estimate noise intensity directly from the raw perturbed input $x+\epsilon$. Instead, we adopt a probabilistic representation that separates semantic content and noise characteristics. Specifically, each modality input is encoded into a Gaussian-distributed representation:
\begin{equation}
z \sim \mathcal{N}(\mu( \varphi_{1}(\boldsymbol{\theta}_{1},{x_i}^{m_1} )), \Sigma( \varphi_{1}(\boldsymbol{\theta}_{1},{x_i}^{m_1} )),
\end{equation}
where $\mu(.)$ captures semantic information and $\Sigma(.)$ models uncertainty. Here, the noise estimator receives the variance term rather than the raw signal:
\begin{equation}
\mathcal{L}_{\mathrm{est}} 
=\big\|  E(\Sigma( \varphi_{1}(\boldsymbol{\theta}_{1},{x_i}^{m_1} +\epsilon))-\sigma_k\big\|_2^2.
\end{equation}
This design ensures that the estimator learns to infer noise intensity from intrinsic distributional uncertainty instead of relying on low-level pixel distortions, thereby improving generalization to diverse noise patterns.

Then the inference uncertainty is defined as follows:
\begin{equation}
    {\rho_i}^{m_1} = E^2(\Sigma( \varphi_{1}(\boldsymbol{\theta}_{1},{x_i}^{m_1})).
\end{equation}

In practice, the estimator $E(\cdot)$ is implemented as a lightweight two-layer MLP operating, which introduces negligible computational overhead while providing strong robustness and noise-level calibration capability.

\subsubsection{Modality-dependency Calculator}
To quantify each modality's contribution to the final prediction, the modality-dependency calculator measures the sensitivity of the model's prediction to the removal of each modality. The core idea is that if dropping a modality causes a larger change in the prediction, then the model relies more on this modality.

Specifically, let $\pi^{\tau}$ denotes the fused logit output from all modalities, and $\pi^{m_2}$ ($\pi^{m_1}$) is the logit obtained when modality $m_1$ ($m_2$) is dropped. The dependency scores $d^{m_1}$ and $d^{m_2}$ that quantify the contribution of each modality to the final prediction are computed as:

\begin{equation}
d^{m_1} = \|\pi^{\tau} -\pi^{m_2} \|_1, \quad
d^{m_2} = \|\pi^{\tau} - \pi^{m_1} \|_1.
\end{equation}
These scores are then normalized to produce scalar modality weights:
\begin{equation}
\alpha^{m_1} = M*\frac{d^{m_1}}{d^{m_1}+d^{m_2}}, \quad
\alpha^{m_2} = M*\frac{d^{m_2}}{d^{m_1}+d^{m_2}},
\end{equation}
where M=2, meaning the number of modalities. $\alpha^{m_1}$ and $\alpha^{m_2}$ reflect the relative importance of each modality and can be used to reweight modality features in the fusion module adaptively. This mechanism allows the model to mitigate dual-suppression effects caused by initial dependency bias and uncertainty-based weighting, ensuring a more balanced and robust multimodal integration.

Notably, modality dropout variants reuse the same fusion and classification modules as the full multimodal branch. No additional parameters or network copies are created. Therefore, the modality-dependency computation introduces negligible overhead.

\subsubsection{Progressive Optimization Strategy}

To effectively train the UDML framework without inducing gradient conflicts between multiple objectives, we introduce a progressive optimization strategy. The core idea is to gradually incorporate different learning targets while preserving the stability of the primary task. Concretely, the training proceeds in two stages: multimodal representation pre-training and noise-aware training.

 \textbf{Stage 1: Multimodal representation pre-training.}  The model is first trained on clean, unperturbed data to optimize the main task (e.g., classification) and learn stable multimodal representations. The loss in this stage is defined as follows: 
\begin{equation}
\mathcal{L}_{\mathrm{total}} = \mathcal{L}_{\mathrm{task}} + \mathcal{L}_{\mathrm{uni}},
\end{equation}
where $ \mathcal{L}_{\mathrm{task}} =\mathcal{L}_{CE}(f(x_i),y)$ denotes the multimodal loss for the task at hands; $ \mathcal{L}_{\mathrm{uni}} =\mathcal{L}_{CE}(f({x_i}^{m_1}),y) + \mathcal{L}_{CE}(f({x_i}^{m_2}),y)$ denote the unimodal loss for modality-dependency calculator.
    
\textbf{Stage 2: Noise-aware training.}     After the primary encoder is stabilized, controlled perturbations are introduced to the inputs for training the  noise-aware uncertainty estimator. The loss in this stage is defined as follows:
    \begin{equation}
\mathcal{L}_{\mathrm{total}} = \mathcal{L}_{\mathrm{task}}+ \mathcal{L}_{\mathrm{uni}}+\mathcal{L}_{\mathrm{est}} .
\end{equation}
Importantly, the gradient from the noise estimation loss is blocked from backpropagating into the modality encoders, preventing conflicts with the main task optimization and avoiding performance degradation. This progressive scheme allows the UDML framework to jointly learn multimodal representations, noise estimation, and the primary task within a standard training schedule while ensuring stable convergence and robust task performance.

\begin{table*}[t]
        \centering
        \caption{Performance comparison of different methods on multimodal classification tasks on MVSA-Single,  CREMA-D, and Kinetics-Sounds datasets.}
        \label{tab:performance}
        \setlength{\tabcolsep}{4mm}{
        \begin{tabular}{c|cc|cc|cc}
            \hline
            \multirow{2}{*}{Methods} & \multicolumn{2}{c|}{MVSA-Single} & \multicolumn{2}{c|}{ CREMA-D} & \multicolumn{2}{c}{Kinetics-Sounds} \\
            \cline{2-7}
            & Acc & F1 & Acc & F1 & Acc & F1 \\
            \hline
            Late Fusion\cite{late1} & 76.88 & 75.72 & 58.83 & 59.43 & 64.97 & 65.21 \\
            MMBT\cite{mmbt}  & 78.50 & - & 64.25 & 65.43 & 65.89 & - \\
            TMC\cite{tmc} & 76.06 & 74.55 & 65.43 & 66.10 & 66.12 & 66.51 \\
            ITIN\cite{itin} & 75.19 & 74.97 & - & - & - & - \\
            MVCN\cite{mvcn} & 76.06 & 74.55 & - & - & - & - \\
            QMF\cite{qmf} & 78.07 & 77.18 & 66.92 & 67.02 & 66.81 & 67.25 \\
            EAU\cite{eau} & 79.15 & 78.36 & 67.61 & 67.90& 68.05& 68.36 \\
            LDDU \cite{ldd} & 79.71 & 79.12 & 68.56 & 68.77& 68.95& 69.27 \\ \hline
            UDML & \textbf{80.79}&  \textbf{80.10}  & \textbf{70.02} & \textbf{70.58} & \textbf{69.98} & \textbf{70.26} \\
            \hline
        \end{tabular}}
    \end{table*}

\begin{table*}[ht]
    \centering
    \caption{Performance comparison of different methods for multimodal regression tasks on CMU-MOSI and CMU-MOSEI datasets.}
    \label{tab:mosi}
    \setlength{\tabcolsep}{4mm}{
    \begin{tabular}{c|ccc|ccc}
        \hline
        \multirow{2}{*}{Methods} & \multicolumn{3}{c}{CMU-MOSI} & \multicolumn{3}{c}{CMU-MOSEI} \\
        \cline{2-7}
        & Acc7 & F1 & Corr & Acc7 & F1 & Corr \\
        \hline
        MIB\cite{mib}   & 48.6 & 85.3 & 0.798 & 54.1 & 86.2 & 0.790 \\
        HMA\cite{hma} & 45.3 & 85.6 & 0.782 & 52.8 & 85.4 & 0.787 \\
        MIM\cite{mim}   & 47.0 & 85.9 & 0.805 & 52.5 & 86.3 & 0.792 \\
        GCNet\cite{gcnet}  & 44.9 & 85.1 & - & 51.5 & 85.2 & - \\
        ConFEDE\cite{ConFEDE}  & 42.3 & 85.5 & 0.784 & 54.9 & 85.8 & 0.780 \\
        DiCMoR\cite{DiC-MoR}   & 45.3 & 85.6 & - & 53.4 & 85.1 & - \\
        DMD\cite{dmd}  & 45.6 & 86.0 & - & 54.5 & 86.6 & - \\
        EAU\cite{eau} & 48.8 & 86.2 & 0.809 & 54.8 &86.9& 0.816 \\ 
        LDDU \cite{ldd} & 49.2 & 86.4 & 0.814 & 55.2& 87.1& 0.821 \\ \hline
        UDML & \textbf{50.3} & \textbf{86.8} &\textbf{0.825} & \textbf{56.3} & \textbf{88.0} & \textbf{0.832} \\ 
        \hline
    \end{tabular}}	
		\end{table*}

\section{Experiments}
\subsection{Experimental Settings}
\textbf{Datasets}. We conduct extensive evaluations across five multimodal benchmarks with diverse modality combinations. Specifically, CMU-MOSI~\cite{mosi} and CMU-MOSEI~\cite{mosei} are tri-modal sentiment analysis datasets containing text, visual, and audio modalities. MVSA-Single~\cite{mvse} is a bi-modal image–text dataset for multimodal sentiment analysis. CREMA-D~\cite{Crema} and Kinetics-Sounds (KS)~\cite{ks} provide bi-modal audio–visual data for emotion recognition and action understanding, respectively. This diverse set of datasets allows us to comprehensively validate our method under heterogeneous modality configurations.

\textbf{Implementation Details}. For the tri-modal video datasets CMU-MOSI and CMU-MOSEI, we adopt the same feature extractors as prior dynamic multimodal learning works~\cite{qmf,eau}: FACET for visual, COVAREP for acoustic, and BERT for textual modalities. For the image-text dataset MVSA-Single, we also adopt the same feature extractors as prior works~\cite{qmf,eau}, ResNet-152 for RGB images, and BERT for text inputs. All models are optimized using Adam with an initial learning rate of $1 \times 10^{-5}$, a mini-batch size of 16, and we apply a Reduce-on-Plateau scheduler for learning rate adjustment.

For the CREMA-D and Kinetics-Sounds datasets, we employ ResNet18 as the backbone encoder, consistent with prior works~\cite{ogm,pmr}. For the CREMA-D dataset, a single frame is selected from each video clip and resized to 224×244 to serve as the visual input, while the corresponding audio is converted into a spectrogram of size 257×299 using librosa~\cite{librosa}. In the case of the Kinetics-Sounds dataset, three frames are uniformly sampled from each video clip and resized to 224×224 for visual input. The entire audio data is transformed into a spectrogram with dimensions of 257×1,004. Training is carried out with configurations consistent with previous works~\cite{ogm,pmr}, including a mini-batch size of 64, an SGD optimizer with momentum set to 0.9, a learning rate of 1e-3, and a weight decay of 1e-4.

\textbf{Evaluation Metric}. For the classification tasks on MVSA-Single, CREMA-D, and Kinetics-Sounds, we report accuracy (Acc) and F1 score. For the regression-style sentiment analysis tasks on CMU-MOSI and CMU-MOSEI, we follow prior work~\cite{qmf,eau} and report 7-class accuracy (Acc7), F1 score, and Pearson correlation coefficient (Corr).


\textbf{Comparison settings}.
We compared UDML with existing dynamical multimodal learning methods on the five datasets, including MIB~\cite{mib}, HMA~\cite{hma}, MIM~\cite{mim}, GCNet~\cite{gcnet}, ConFEDE~\cite{ConFEDE}, DiC-MoR~\cite{DiC-MoR}, DMD~\cite{dmd}, and QMF~\cite{qmf}. Moreover, we also make fair comparisons with the recent SOTA methods, including EAU~\cite{eau} and LDDU~\cite{ldd}. Particularly, similar to the counterpart method EAU~\cite{eau}, we evaluate our method on noisy datasets to observe the model robustness.

Notably, despite UDML needing a two-stage optimization process, the total number of training epochs is consistent with that of existing methods, ensuring a fair comparison. Specifically, Stage 1 lasts for half of the total epochs, while Stage 2  lasts for the remaining half.

\begin{table*}[ht]
    \centering
    \caption{Performance comparison of different methods when 50\% of the modalities suffer from salt and Gaussian noise, respectively. The mean and variance of the noise are 0 and $\epsilon$, respectively. }
    \label{multi-dml}
    \begin{tabular}{c|c|c|cc|cc}
        \hline
        \multirow{2}{*}{Datasets}     & \multirow{2}{*}{Methods} & Clean          & \multicolumn{2}{c|}{Salt}               & \multicolumn{2}{c}{Gaussian}                  \\ \cline{3-7} 
        &                         & Acc@$\epsilon $=0       & \multicolumn{1}{c|}{Acc@$\epsilon $=5}       & Acc@$\epsilon $=10      & \multicolumn{1}{c|}{Acc@$\epsilon $=5}       & Acc@$\epsilon $=10      \\ \hline

        \multirow{8}{*}{MVSA-Single} & Late fusion \cite{late1}                    & 76.88          & \multicolumn{1}{c|}{67.88}          & 55.43          & \multicolumn{1}{c|}{63.46}          & 55.16          \\ 
        & MMBT\cite{mmbt}                    & 78.50          & \multicolumn{1}{c|}{74.07}          & 51.26          & \multicolumn{1}{c|}{71.99}          & 55.35          \\ 
        & TMC\cite{tmc}                     & 74.88          & \multicolumn{1}{c|}{68.02}          & 56.62          & \multicolumn{1}{c|}{66.72}          & 60.36          \\ 
        & QMF \cite{qmf}                    & 78.07          & \multicolumn{1}{c|}{73.90}          & 60.41          & \multicolumn{1}{c|}{73.85}          & 61.28          \\ 
        & EAU  \cite{eau}                   & 79.15          & \multicolumn{1}{c|}{74.81}          & 61.04          & \multicolumn{1}{c|}{73.89}          & 62.04          \\ 
        & LDDU  \cite{ldd}                   & 79.71          & \multicolumn{1}{c|}{75.36}          & 61.36          & \multicolumn{1}{c|}{74.10}          & 62.25          \\         
        \cline{2-7} 
        & UDML                    & \textbf{80.79} & \multicolumn{1}{c|}{\textbf{78.26}} & \textbf{64.52} & \multicolumn{1}{c|}{\textbf{77.87}} & \textbf{63.62} \\ \hline

        \multirow{8}{*}{CREMA-D}     & Late fusion\cite{late1}             & 58.83          & \multicolumn{1}{c|}{50.99}          & 48.75          & \multicolumn{1}{c|}{49.49}          & 47.00          \\ 
        & MMBT\cite{mmbt}                    & 64.25          & \multicolumn{1}{c|}{55.27}          & 52.98          & \multicolumn{1}{c|}{54.32}          & 51.75          \\ 
        & TMC\cite{tmc}                     & 65.43          & \multicolumn{1}{c|}{58.86}          & 54.22          & \multicolumn{1}{c|}{56.93}          & 53.37          \\ 
        & QMF\cite{qmf}                     & 66.92          & \multicolumn{1}{c|}{59.56}          & 55.45          & \multicolumn{1}{c|}{58.03}          & 54.21          \\ 
        & EAU\cite{eau}                     & 67.61          & \multicolumn{1}{c|}{60.02}          & 56.33          & \multicolumn{1}{c|}{59.59}          & 55.57  \\ 
         & LDDU\cite{ldd}                     & 68.56          & \multicolumn{1}{c|}{61.25}          & 57.12          & \multicolumn{1}{c|}{60.32}          & 56.14  \\        
        
        \cline{2-7}
        & UDML                    & \textbf{70.02} & \multicolumn{1}{c|}{\textbf{68.32}} & \textbf{64.48} & \multicolumn{1}{c|}{\textbf{67.53}} & \textbf{62.96} \\ \hline

    \end{tabular}
\end{table*}

\subsection{Experimental Results}
\textbf{Performance and comparison on multimodal classification tasks}. Table~\ref{tab:performance} reports the multimodal recognition results of our proposed UDML method compared to a range of baselines on the MVSA-Single,  CREMA-D, and Kinetics-Sounds datasets. As shown, UDML consistently achieves the best performance across all datasets, outperforming both static fusion methods and dynamic fusion approaches. Particularly, compared with the recent state-of-the-art method LDDU, UDML achieves at least 1\% absolute improvements on all datasets. These results highlight UDML's effectiveness in dynamic multimodal fusion.

\textbf{Performance and comparison on multimodal regression tasks}. Table~\ref{tab:mosi} presents the results of UDML and its competitors on the CMU-MOSI and CMU-MOSEI datasets for multimodal regression tasks. UDML also achieves the best performance across all metrics and both datasets.  Specifically,  On CMU-MOSI, UDML obtains 50.3\% Acc7, 86.8\% F1, and 0.825 Corr, outperforming prior representative methods such as EAU and LDDU. Similarly, on CMU-MOSEI, UDML achieves 56.3\% Acc7, 88.0\% F1, and 0.832 Corr, marking a consistent improvement over existing approaches.  Notably, UDML maintains strong performance on both small-scale (MOSI) and large-scale (MOSEI) datasets, demonstrating robustness and generalization in multimodal sentiment regression.

\textbf{Performance and comparison on noisy multimodal datasets}. 
To assess the effectiveness of UDML  in handling data noise, we follow QMF~\cite{qmf} and conduct more evaluation when 50\% of the multimodal samples are corrupted by salt or Gaussian noise with different intensities. As shown in Table~\ref{multi-dml}, performance degrades across all methods as noise intensity increases. However, UDML consistently achieves the best performance across all noise settings on both MVSA-Single and CREMA-D. Compared with existing dynamic fusion methods (e.g., EAU, LDDU), UDML shows significantly smaller performance degradation as noise intensity increases, demonstrating strong robustness to modality corruption. This improvement stems from the  noise-aware uncertainty estimator providing calibrated and noise-agnostic uncertainty estimation, which enables reliable down-weighting of heavily corrupted modalities.

Importantly, although the noise estimator is trained using only Gaussian perturbations, UDML generalizes robustly to salt noise as well, confirming that the estimator does not overfit to noise type and instead learns a noise-agnostic measure of corruption. This cross-noise generalization ability indicates that the proposed variance-based uncertainty modeling successfully captures universal corruption cues shared across different noise distributions.

Moreover, the improvement is particularly significant on CREMA-D. This is because CREMA-D exhibits a large modality bias where the audio modality dominates the optimization and has a higher contribution to the logit output. Conventional dynamic fusion strategies further amplify this bias when noise is present, leading to dual suppression of the visual modality. In contrast, our dependency realignment mechanism explicitly counteracts such imbalance by correcting the modality contribution shift, ensuring that both modalities remain effectively utilized. As a result, UDML achieves more stable and superior multimodal collaboration, especially in scenarios with strong modality asymmetry





\subsection{Ablation Study}


\begin{table}[]
\centering
\caption{Ablation of the UDML method when 50\% of the modalities suffer from Gaussian noise ($\epsilon=5$). We report the result of complete UDML (Full), the UDML without noise-aware uncertainty estimator (-NUE),  the UDML without Modality-dependency Calculator (-MC), and the UDML without Progressive Optimization Strategy(POS).}
\begin{tabular}{c|c|c|c}
\hline
Dataset & MVSA &  CREMA-D & Kinetics-Sounds \\ \hline
Settings & Acc & Acc & Acc \\ \hline
Full & \textbf{77.87} & \textbf{67.53} & \textbf{66.36} \\
- NUE & 74.56 & 64.60 & 64.33 \\
- MC & 75.44 & 64.10 & 65.41 \\
- POS &76.16 & 66.49 & 64.92 \\ \hline
\end{tabular}
\end{table}

\textbf{The ablation study of UDML}. We conduct experiments to study the impact of the  noise-aware uncertainty estimator (NUE), the Modality-dependency Calculator (MC), and the Progressive Optimization Strategy (POS)  on the performance of UDML. As shown in Table 4, removing any component leads to performance degradation, confirming the necessity of each part. Excluding the noise-aware uncertainty estimator (–NUE) notably reduces accuracy (e.g., 77.87→74.56 on MVSA-Single), indicating that heuristic uncertainty estimation fails to correctly assess noise levels. Removing the Modality-dependency Calculator (–MC) causes the most drop, verifying that addressing modality bias is crucial to avoid dual suppression of difficult modalities. Meanwhile, removing the Progressive Optimization Strategy (–POS) also hurts performance due to gradient conflict during joint optimization. Overall, the full UDML achieves the best performance, demonstrating that accurate noise estimation, dependency realignment, and stable optimization jointly contribute to unbiased dynamic multimodal learning.

\begin{table}[]
\centering
\caption{ Performance evaluation after integrating the UDML method into CNN-based and Transformer-based architectures. † indicates that the UDML is applied.}
\label{gene}
\begin{tabular}{c|c|c|c}
\hline
Datasets & MVSA &  CREMA-D & Kinetics-Sounds \\ \hline
Structure & Acc & Acc & Acc \\ \hline
MMTM & 78.86 & 51.86 & 56.70 \\
MMTM$^\dagger$ & \textbf{85.68} & \textbf{56.88} & \textbf{62.72} \\ \hline
mmFormer & 79.69 & 59.69 & 64.72 \\
mmFormer$^\dagger$ & \textbf{85.33} & \textbf{63.14} & \textbf{67.51} \\ \hline
\end{tabular}
\end{table}
\textbf{Generalization to different multimodal architectures}. To demonstrate the generality of UDML in various settings, we integrate it with two representative intermediate fusion methods, MMTM~\cite{mmtm} and mmFormer~\cite{mmformer}, and evaluate the performance across multiple datasets. MMTM is a CNN-based architecture that fuses multimodal intermediate features using squeeze-and-excitation operations, while mmFormer is a Transformer-based model employing cross-attention for feature fusion. Specifically, UDML is inserted before the fusion block to regulate the reweighted multimodal representation. The training schedules and data augmentation strategies are kept identical to the respective baseline implementations to guarantee a fair comparison.

As shown in the right part of Table~\ref{gene}, although MMTM and mmFormer already achieve competitive performance without UDML,  the introduction of UDML also brings significant performance gains. Specifically, on the MVSA-Single dataset, the performance of MMTM improves from 78.86\% to 85.68\% when the UDML is applied. These results confirm the robustness and versatility of UDML across different architectures.


\begin{figure}[t]
	\centering
	\includegraphics[width=0.5\textwidth]{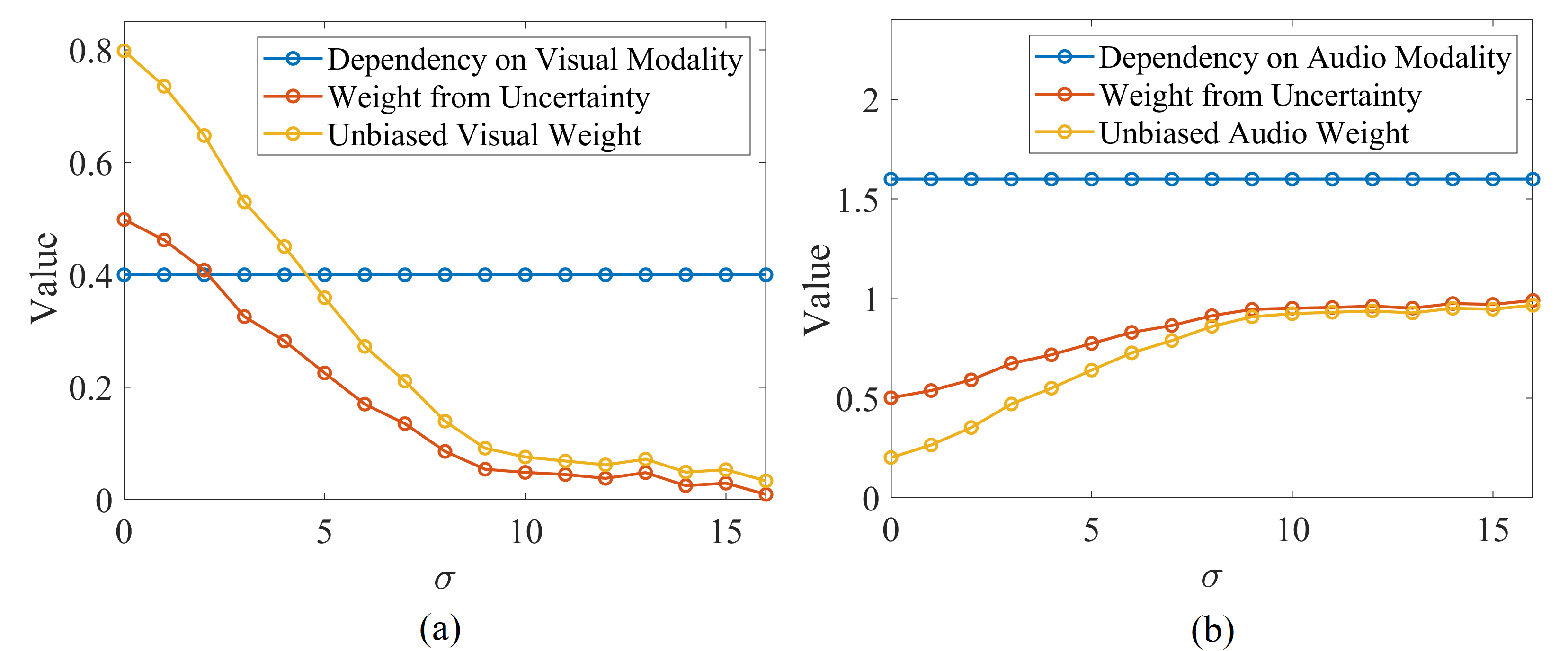}
    
	\caption{Visualization of dynamic multimodal fusion for audio-visual classification on the CREMA-D dataset as varying levels of noise ($\sigma$) are injected into the visual modality.}
	\label{visual}
\end{figure}



\textbf{Visualization}. To understand the mechanism of UDML intuitively, we visualize the inference coefficients of each modality on the CREMA-D dataset when different noises are added to the image modality. As shown in Fig~\ref{visual}(a), the weight for visual modality (red curve) smoothly decreases as the visual noise increases, and eventually approaches zero when the noise is large. This demonstrates that the proposed noise-aware uncertainty estimator can accurately distinguish both low-noise and high-noise cases, overcoming the misestimation problem of traditional uncertainty estimation methods.


Meanwhile, we can see that the model assigns a stronger intrinsic dependency to the audio modality (=1.6) than to the visual modality (=0.4), regardless of noise levels. This demonstrates the dependency bias of the multimodal model. By integrating both the noise-aware weighting and the modality dependency, the final unbiased fusion coefficient corrects for this reliance bias. Specifically, UDML amplifies the visual weight by 1.6× and suppresses the audio weight by 0.4×. For example, the final weights become approximately 0.8:0.2 (visual: audio) instead of 0.5:0.5 when the noise is 0.  This ensures that fusion respects the intrinsic discriminative contributions while remaining robust to noise, achieving a balanced and reliable multimodal integration.


\section{Conclusion}


In this paper, we identify two fundamental challenges in multimodal fusion: unreliable modality-uncertainty estimation under extreme noise and the dual suppression effect, in which low-contribution modalities are further down-weighted by high uncertainty. As a result, dynamic fusion may even underperform static fusion. To address this, we propose UDML, a dynamic multimodal learning framework that combines reliable uncertainty estimation with dependency-aware fusion. Visual analysis shows that our uncertainty estimator accurately tracks modality degradation and enables adaptive suppression of corrupted modalities. Meanwhile, the derived reliance coefficients expose intrinsic contribution imbalance, which is effectively corrected by our unbiased weighting strategy. Extensive experiments on multiple multimodal datasets demonstrate that UDML yields stronger robustness, more stable fusion behavior, and better overall performance than existing fusion methods.

\section{Limitation}
UDML mainly focuses on modality-level bias and does not explicitly address sample-level bias. For example, tail-class samples may present high uncertainty because of limited training data rather than modality noise. In such cases, the model may over-suppress these samples by assigning unnecessarily low modality weights. This suggests that uncertainty in multimodal learning may arise not only from modality corruption but also from data imbalance and sample difficulty. In future work, we plan to investigate how to disentangle these sources of uncertainty and integrate sample-level bias correction into dynamic multimodal fusion.

\section*{Acknowledgments}
This work is supported by Ministry of Science and Technology of Sichuan Province Program (2026NSFSC0430).

{
    \small
    \bibliographystyle{ieeenat_fullname}
    \bibliography{main}
}


\end{document}